\documentclass[lettersize,journal]{IEEEtran}
\usepackage{amsmath,amsfonts}
\usepackage{algorithmic}
\usepackage{algorithm}
\usepackage{array}
\usepackage[caption=false,font=normalsize,labelfont=sf,textfont=sf]{subfig}
\usepackage{textcomp}
\usepackage{color}
\usepackage{stfloats}
\usepackage{url}
\usepackage{cite}
\usepackage{verbatim}
\usepackage{graphicx}
\usepackage{amsthm}
\begin{document}

\title{A Novel Knowledge-Based Genetic Algorithm for Robot Path Planning in Complex Environments} 

\author{Yanrong~Hu,~
        and~Simon~X.~Yang,~\IEEEmembership{Senior Member,~IEEE}
\thanks{
This work was supported by Natural Sciences and
Engineering Research Council (NSERC) of Canada.}
\thanks{The authors are with the Advanced Robotics and Intelligent Systems
(ARIS) Lab, School of Engineering, University of Guelph,
Guelph, Ontario, N1G 2W1, Canada 
(e-mail: \{huy; syang\}@uoguelph.ca). }
\thanks{Corresponding author: Simon X. Yang }}


\maketitle

\begin{abstract}
In this paper, a novel knowledge-based genetic algorithm for path planning of a mobile robot in unstructured complex environments is proposed, where five problem-specific operators are developed for efficient robot path planning.  The proposed genetic algorithm incorporates the domain knowledge of robot path planning into its specialized operators, some of which also combine a local search technique.  A unique and simple representation of the robot path is proposed and a simple but effective path evaluation method is developed, where the collisions can be accurately detected and the quality of a robot path is well reflected.  The proposed algorithm is capable of finding a near-optimal robot path in both static and dynamic complex environments.  The effectiveness and efficiency of the proposed algorithm are demonstrated by simulation studies.  The irreplaceable role of the specialized genetic operators in the proposed genetic algorithm for solving the robot path planning problem is demonstrated through a comparison study.
\end{abstract}


\section{Introduction}

Path planning is a fundamentally important issue in mobile robotics. 
A mobile robot needs to carry out a series of tasks in order to navigate in an
environment. First, the mobile robot needs to sense the environment to obtain
knowledge about the target and obstacles and to localize itself in the
environment. 
Then a path planner is needed to construct a suitable collision-free path
for the robot to move from the start location to the target location.
Very often it is desirable that the robot path is 
optimal or near-optimal with respect to time, distance or energy. 
Distance is a commonly adopted criterion. After a
suitable path is available, a path tracking algorithm makes the mobile robot
to follow the path and reach the target. Although each task involved in mobile
robot navigation has different challenges, when the
environment information is already available, path planning becomes the
primary task for the mobile robot. This paper assumes that
the environment information is completely known in static environments and
the new environment information is already updated in dynamic environments.
Therefore, it is basically path planning for global navigation. The algorithm deals with
path planning problem only, and does not involve localization and path tracking.
When the environment information is partially known or only short-range information
of the environment can be obtained during robot's navigation, it is called 
local navigation. Path planning for local navigation usually cannot be separated
from other tasks such as localization and path tracking, and algorithms for navigation
instead of just path planning are required. Markov Decision Process (MDP) is widely
used for this task
\cite{Theocharous&Mahadevan:2002,Laroche&etal:1999,Laroche:2000,Cassandra&etal:1996}. 
MDP is mainly used for indoor robot's navigation, especially for 
corridor or office environments. MDP-based approaches are able to deal with some uncertainty, 
which makes it can be used in realistic applications. However, MDP also usually requires  
the structure of the environment to be known, and it usually deals with environments with
simple (rectangular or block) obstacles. Besides, the path planned by MDP approaches are
usually not intended to be optimal, but feasible. 
The path planning in this paper falls out of the 
catagory of MDP-based path planning. It emphasises on its ability to plan optimal or 
near optimal path among complicated obstalces.

Robot Path planning has been an active research area, and 
many methods have been developed to tackle this problem 
\cite{Latombe:1991}, such as
global configuration-space methods \cite{Lozano-Perez:1983,Sharir:1989}, 
potential field methods \cite{Khosla&Volpe:1988,Rimon&Doditschek:1992}, 
and neural networks based approaches
\cite{Yang&Meng:2000a,Yang&Meng:2000b,Yang&Meng:tsmc2001,Yang&Meng:tnninpress}.  
Each method has its own strength over others on certain aspects. 
Generally, the main difficulties for path planning problem are 
computational complexity, local optimum, and
adaptability. Researchers have always been seeking alternative and more
efficient ways to solve the problem.

There is no doubt that path planning can be viewed as an optimization problem
(e.g., the shortest distance) under certain constraints (e.g., 
obstacle avoidance).  
Since the appearance of genetic algorithms (GAs) around before 1975 
\cite{Holland:1975}, GAs have been used for solving 
various optimization problems successfully, 
particularly very complex problems where conventional searching approaches
would fail.  
GAs are advanced stochastic search techniques analogous to natural evolution 
based on the principle of ``survival of the fittest'' 
\cite{Davis:1987,Michalewicz:1994}. Potential solutions of a problem
are encoded as chromosomes, which form a population. Each individual of the
population is evaluated by a fitness function. A selection mechanism based on
the fitness is applied to the population and the individuals strive for
survival.  The fitter ones have more chance to be selected and to reproduce
offspring by means of genetic transformations such as crossover and mutation.
The process is repeated and the population is evolved generation by
generation. After many generations, the population converges to solutions of
good quality, and the best individual has a good chance to be the optimal or
near-optimal solution. The feature of parallel search and the ability of
quickly locating high performance region \cite{Davis:1987} contribute to the
success of GAs on many applications. 

It is not surprising that many researchers have applied GAs to 
path planning of mobile robots (e.g., 
\cite{Sugihara&Smith:1997a,Sugihara&Smith:1997b,
Sugihara&Smith:1997c,Ashiru&etal:1996,Gallardo&etal:1998}). 
However, like most early GA applications, 
most of those methods adopt classical GAs that use
fixed-length binary strings and two basic genetic operators (crossover and
mutation), while few modifications were made to the algorithms. 
Sugihara {\em at al.} 
\cite{Sugihara&Smith:1997a,Sugihara&Smith:1997b, Sugihara&Smith:1997c}
proposed a genetic algorithm for robot path planning 
with fix-length binary string
chromosomes based on cell representation of the mobile robot environment.  A
workspace is divided into grids, in which a mobile robot can only move from
one cell to the adjacent one. The grids occupied by obstacles are assigned
with more weight so that a path intersecting with an obstacle has more cost.
At each cell, there are eight possible directions to move. A path is
represented by a sequence of moving directions starting from the start point.
In a chromosome, a gene is a relative direction, and what it represents can
only be interpreted from all the previous genes.  Since the number of moving 
steps to reach the target is uncertain, variable-length chromosome comes
very naturally. However, in order to 
use the fixed-length chromosomes and the basic genetic operators, 
the robot path is limited to be $x$-monotone or $y$-monotone, which means
that the projection of the path on $x$-axis or $y$-axis is non-decreasing.
Obviously, this monotone approach puts some restriction on path-planning
solutions and makes it unsuitable for complex environments. 
Another disadvantage of the fixed-length GA approach is its biased encoding.  
The entire path is represented by a sequence of relative directions 
and distances.  A small
change at the beginning of a binary string may dramatically affect the entire
path while changes at the end part of the binary string have only minor
effects.  Moreover, besides the infeasible robot path that intersects with
obstacles, this encoding suffers from the problem of other invalid paths: 
after crossover or mutation the new path may go beyond the environment; 
and there is no guarantee that a path reaches the target 
because the generated sequence of directions may not necessarily lead 
the robot path to reach the target.  
With too many invalid chromosomes, the performance of the genetic algorithm 
would be highly affected.  The approach of using standard GAs with 
an inefficient binary encoding and by limiting the path planning 
seems not very effective. 
Although Tu and Yang \cite{Tu&Yang:2003} improved this approach by  
using variable-length chromosomes instead of fixed-length chromosomes,
the efficiency of the approach is still not improved and it would take hours to 
evolve a solution because of the biased and inefficient encoding.

The classical GAs use binary strings and two basic genetic operators. After
encoding solutions to a problem, the classical GAs are more like ``blind"
search. They perform well when no or very little prior knowledge is available.
However, GAs do not have to be ``blind" search when additional knowledge
about the problem is available. The available knowledge could be incorporated 
into GAs to improve the efficiency of GAs 
\cite{Michalewicz:1994,Grefenstette:1987}. There are many ways to 
incorporate additional knowledge into GAs. First, problem-specific
knowledge can be used in a more natural way of chromosome representation of
potential solutions to a problem instead of binary strings.  Second, genetic
operators can be modified to use the knowledge in the operation 
and to fit the problem more. In addition, heuristic methods (e.g.
local search) can be combined into the genetic operators. Furthermore, 
problem-specific knowledge can be used to guide construction of 
initial population.
One or more of the above techniques were experienced by some researchers in
various applications, and obtained satisfactory results 
\cite{Blum&Eskandarian:2002,Janikow&Michalewicz:1990,Suh&Gucht:1987,
Akbarzadeh-T&Jamshidi:1997,Janikow:1993,Tripathi&etal:1996,Ramsey&Grefenstette:1993,Andres&etal:2000,Louis&Zhao:1995,Antonisse&Keller:1987,Davis&Coombs:1987,Coombs&Davis:1987}.

Robot path planning is such a complex problem that 
the incorporation of domain knowledge into the GAs 
would significantly improve the efficiency in obtaining the solution.
Graph technique is a traditional way of representing
the environment where a mobile robot moves around in a workspace. 
Shibata {\em et al.}
\cite{Shibata&Fukuda:1993,Shibata&etal:1992} proposed a genetic algorithm
based on MAKLINK graph environment representation \cite{Habib&Asama:1991}.
The MAKLINK uses a free-link concept to construct the available free space
between obstacles within an environment in terms of free convex areas. The
free-link is a line connecting the corners of two polygonal obstacles (the
working space boundary is treated as obstacles too), and should not
intersect with any of the edges of the obstacles. Free-convex area is formed
by free-links and edges of obstacles. The midpoint of each free-link is used
as a node in the graph. Each node is numbered and a path is encoded as a
sequence of these nodes with variable length. To make sure the path is valid
(collision free), adjacent genes must be the numbers that are connected
with a free-link in the graph. In this genetic algorithm, orderly based and 
variable-length chromosomes are used, which is much more natural than 
fix-length binary chromosomes. 
This graph-based method needs to form a configuration space before applying
the genetic algorithm. The forming of the configuration space could be very
time consuming, and it is only suitable for static environments because a
slight change of the environment would cause the re-computation of the graph.
Besides, a robot path can only visit the fixed nodes pre-determined 
by the graph, not everywhere in the workspace. 
Such a robot path can only be sub-optimal and
lack flexibility. One advantage of this method is that every
possible path is feasible because the graph is generated under the
consideration of collision avoidance.  
This makes the evaluation of a chromosome relatively simple and fast.

More effective path-planning methods can be found in
\cite{Hocaoglu&Sanderson:2001} and
\cite{Xiao&etal:1997,Trojanowski&etal:1997,Xiao&etal:1996,Xiao:1997}, which
deviate from standard GAs. In \cite{Hocaoglu&Sanderson:2001}, a multi-path
planning algorithm that can be applied to up to six degrees of freedom was
proposed. The algorithm is based on an iterative multi-resolution path
representation. A path is represented by a hierarchically ordered set of
vectors that define path vertices generated by a modified Gram-Schmidt
orthogonalization process
\cite{Hocaoglu&Sanderson:1995,Hocaoglu&Sanderson:1997}.  A
multiple population steady-state GA with fitness sharing is developed for
multiple robot path planning. It uses minimal representation size criterion and
cluster analysis to formulate evolutionary speculation. Xiao {\em et al.}
\cite{Xiao&etal:1997,Trojanowski&etal:1997,Xiao&etal:1996,Xiao:1997} proposed
an evolutionary path planner for both on-line and off-line planning, 
which uses a continuous coordinate representation and 
eight problem-specific genetic operators. 
This approach uses two different fitness functions to evaluate
feasible paths and infeasible paths. Both approaches above use 
problem-specific chromosome structures and non-standard genetic operators, 
and show better results over the early approaches using standard GAs. 
However, both approaches are relatively complicated on the 
problem representation, solution evaluation, or GA structure. 
In \cite{Hocaoglu&Sanderson:2001}, a binary tree for path
representation is needed, which is obtained by iteratively using 
a special method. Also, the multi-population approach makes its GA structure 
more complex than a simple GA. 
Besides, they only dealt with static environments, although in \cite{Xiao&etal:1997}, a suddenly appeared obstacle is considered. 

In this paper, a novel knowledge-based genetic algorithm is proposed. 
It incorporates problem-specific knowledge into many aspects of the algorithm:
encoding, evaluation, and genetic operator. This is different from the above introduced
GA approaches to path planning that only incorporate additional knowledge into one
element of the GA. It uses a
simple yet effective path representation that combines grids and coordinates
representations. The environment including obstacles and end-points 
(starting point and target point) are represented by their natural 
coordinates in a continuous workspace. 
Grids are only applied to the nodes of paths
 to round off the coordinates of the nodes to integers according to a certain resolution. 
Unlike other grid methods, the grids adopted here do not limit the movement of 
the robot path, but simplify the chromosome structure and genetic operation. 
This approach makes it possible to have one number for each gene 
and to use integer numbers instead of real numbers in chromosomes.  
The proposed GA has five knowledge-based genetic operators.  
Problem-specific genetic operators are not only designed with 
the domain knowledge of robot path planning, 
but also incorporate small-scale local search that 
improves efficiency of the operators.  A relatively simple
but effective evaluation method is applied to both feasible and infeasible
solutions to detect collision and 
reflect solution quality. The collision detection 
developed here is suitable for arbitrarily shaped obstacles. 
No commercial or already-made packages are used. 
The proposed GA is suitable for both static and dynamic
environments. In dynamic environments, the robot is capable of avoiding the 
moving obstacles, while moving toward the target. 
The effectiveness and efficiency of the proposed 
knowledge-based GA for mobile robot path planning are 
demonstrated by simulation and comparison studies.

This paper is organized as follows.  The proposed 
knowledge-based genetic algorithm for path planning for a mobile robot 
is presented in Section II, including the problem representation, 
solution evaluation, and five genetic operators specifically designed for
robot path planning.  
Section III presents the simulation results for both static and 
dynamic environments.  The effectiveness of 
the specially developed genetic operators for robot path planning 
and the complexity of the proposed algorithm are discussed in Section IV. 
Finally, some concluding remarks and future work 
are briefly outlined in Section V. 

\section{The Proposed Knowledge Based GA for Path Planning}

The proposed genetic algorithm features its simple and unique problem
presentation, its effective evaluation method and its knowledge-based genetic
operators specifically designed for robot path planning. 
In this section, the problem presentation is first provided. 
Then the evaluation method is presented. After that, 
five problem-specific genetic operators are given. 
The outline of the proposed knowledge-based GA is finally presented. 

\subsection{Problem Representation}

\begin{figure}[htbp]
\begin{center}
\includegraphics[width=0.45\textwidth]{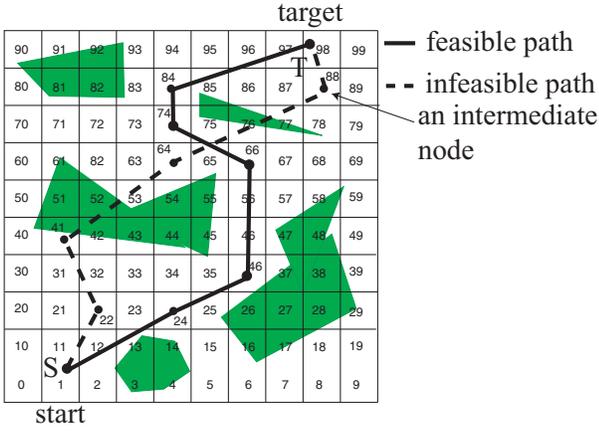}
\caption{\label{f_fig1} 
An example of mobile robot environment and path representation.
Solid line: a feasible path; dashed line: an infeasible path.} 
\end{center}
\end{figure}

Problem representation is a key issue in the applications of GAs.  The
proposed GA uses a simple yet effective path representation.  The mobile
robot environment is represented by a 2-dimensional (2D) 
continuous workspace, where obstacles
are represented by the coordinates of their vertices. The boundary of
obstacles is formed by their actual boundary plus a minimum safety distance
with consideration of the size of the mobile robot, 
which makes it possible
to treat the mobile robot a point in the environment.  
According to a certain resolution, the grids assigned with sequencing integer numbers are 
applied to the workspace. Fig. \ref{f_fig1} gives an example of an environment and 
path representation. Such a grid representation is
different from the one that usually uses grids to discretize the whole
environment and to limit the robot movement to one of its eight adjacent
cells, and also different from the one that 
uses relative directions to represent a path
\cite{Sugihara&Smith:1997a}. The grids here are {\em not} to discretize the whole
environment and do {\em not} affect obstacle representation.  
As shown in Fig. \ref{f_fig1}, the grids have no effects on obstacles, 
start point and target. The grids here are only to form the intermediate nodes of a path. 
A potential robot path
is formed by several line segments connecting the start point $S$,
intermediate nodes, and the target point $T$, where 
$S$ and $T$ are represented by their natural coordinates. 
An intermediate node is a node falling on one
of the grids applied on the workspace and is represented by its associated number.
For instance, in Fig. \ref{f_fig1}, Node 88 is an intermediate node. %

\begin{figure}[htbp]
\begin{center}
\includegraphics[width=7.0cm]{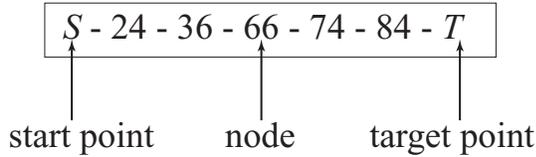}
\caption{\label{f_fig2} 
A sample chromosome: a path represented by nodes falling on grids with different numbers.}
\end{center}
\end{figure}

In fact the grids here 
indicate the resolution that only affects the intermediate nodes, 
and at the same time make it possible to use integers to represent 
a node instead of real-valued coordinates. 
Therefore, the chromosome structure and the
genetic operations are simplified, and thus speed up the computation. Using
grid numbers to represent intermediate nodes is acceptable as long as the
resolution is high enough for the environment in question.  
An example of
path encoding is shown in Fig. \ref{f_fig2}.  A feasible robot path is a
collision-free path, i.e., no nodes fall on any obstacle, and none of the
line segments intersect any obstacles. The length of a chromosome is variable,
between the minimum of $2$ and the maximum length $N_{\mathrm{max}}$. 

\subsection{Evaluation}

\begin{figure}[tbp]
\begin{center}
\includegraphics[width=2in]{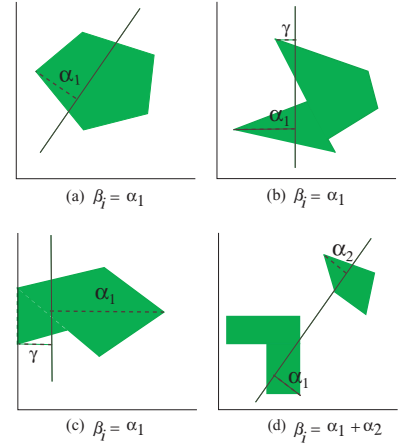}
\caption{Definition of the coefficient $\beta_{i}$.}
\label{f_fig3} 
\end{center}
\end{figure}

A robot path generated by the GA can be either a feasible (collision free) path  
or an infeasible path where at least a line segment intersects an obstacle. 
The evaluation should be able to distinguish feasible and infeasible paths 
and tell the difference of path qualities within either category. In addition, 
it is very important to 
differentiate the qualities of infeasible paths 
because the GA would evolve feasible solutions from those infeasible solutions. 
It requires that those infeasible solutions with better qualities should be easier 
to be evolved by the GA. 
Therefore, the evaluation method should first detect collision. 
Secondly, if the path collides with obstacles, it should tell how deep it intersects, 
i.e., how difficult the path can escape from the obstacles so that 
new solutions can be more-easily evolved from those easier-to-escape solutions. 
If the path is collision free, its quality is simply indicated by the path length.
In this study, the evaluation function is defined as
\begin{equation} \label{eq:fit}
F_{\mathrm{cost}} = \sum_{i=1}^N (d_{i}+ \beta_{i}C),
\end{equation}
where $N$ is the number of line segments of a path,  $d_{i}$ is the Euclidean
distance of the two nodes forming the line segment, and $C$ is a constant. 
Variable $\beta_{i}$ is the coefficient denoting the depth of collision, which is defined as
\begin{equation}
\beta_{i} = \left\{ \begin{array}{ll}
0 & \textrm{if the $i$th line segment is feasible}\\
\sum_{j=1}^M \alpha_{j} & \textrm{if the line segment intersects obstacle(s)} 
\end{array} \right. ,
\end{equation}
where $M$ is the number of obstacles intersecting the line segment, and
$\alpha_{j}$ is determined by considering how deep a line segment intersects
an obstacle $j$.  It is defined as the shortest moving distance for escaping
the intersecting obstacle.  Fig. \ref{f_fig3} explains the definition of
$\alpha_{j}$.  
In Fig. \ref{f_fig3}(a), $\alpha_{1}$ is treated as the
shortest distance to move the line out of the obstacle. 
In Fig. \ref{f_fig3}(b), $\gamma$ is the shortest distance, 
but it is not enough to
move the path away from the obstacle, therefore, instead $\alpha_{1}$ is
assigned to $\beta_{i}$.  
In Fig. \ref{f_fig3}(c), the obstacle is connected
to the wall and is treated as a dead-end.  
The line can only escape from
the other side of the obstacle. Thus $\beta_i$ is assigned as $\alpha_1$.
Fig. \ref{f_fig3}(d) shows the situation that
the line segment intersects two obstacles, it is considered to be 
more difficult for the path to move away from both, 
so the sum of $\alpha_{1}$ and $\alpha_{2}$ is
used for $\beta_i$. 

Accurate collision detection plays an essential role in the evaluation.
Collision detection itself is a hard topic that interests many researchers.
Here an effective algorithm is developed to detect collision between a
straight line and arbitrarily shaped obstacles in a 2D environment. 
The obstacle can be rectangular, convex or concave. 
The method for checking if a line intersects a convex obstacle is simple 
and was described by {\em Pavlidis} \cite{Pavlidis:1982}. 
For an arbitrarily shaped obstacle, collision detection is much more
difficult. 
Some algorithms and packages are available, such as
\cite{Cohen&etal:1995}. Considering the specific requirements needed in the
proposed GA on collision detection between a straight line segment and an
obstacle as well as on quality assessment, we developed an algorithm for 
both collision detection and quality evaluation. 
An arbitrarily shaped obstacle is
treated as a group of convex obstacles connected to each other. The number of
convex obstacles and information regarding connection between them are
obtained. Besides, A Min-Max box of the Group (MMG) is calculated indicating
its minimal and maximal coordinates $x$ and $y$ of the group. Similarly, a
Min-Max box of Obstacle (MMO) is also obtained for each convex obstacle in the
group. If a line does not intersect any MMG, then it is collision free.
Otherwise, check each MMO of the intersected group.  The line intersects the
group if and only if it collides at least one convex obstacle of the group.
Meanwhile, the depth of intersection is calculated with the consideration of
related vertices in the group and connection information.  The use of min-max
box decreases the computation time.

The proposed evaluation method gives a penalty to infeasible paths, 
but still keeps them in the population pool because they might become good
feasible solutions after certain genetic transformations.  Importantly, this
evaluation may allow some overlap between fitnesses of feasible and infeasible
solutions by adjusting $C$. It would be beneficial to
give more chance to some good infeasible solutions that would be easily 
to be evolved to good solutions. 
During the evaluation, some information obtained
by the evaluation needs to be recorded so that later it can be used by
some specialized genetic operators as heuristic knowledge without
re-calculation in order to save computation time. The information includes
feasibility (feasible or infeasible),
number of infeasible line segments, 
and which obstacles intersected by which line segments 
of a path.

\subsection{Genetic Operators}

\begin{figure}[tbp]
\begin{center}
\includegraphics[width=0.45\textwidth]{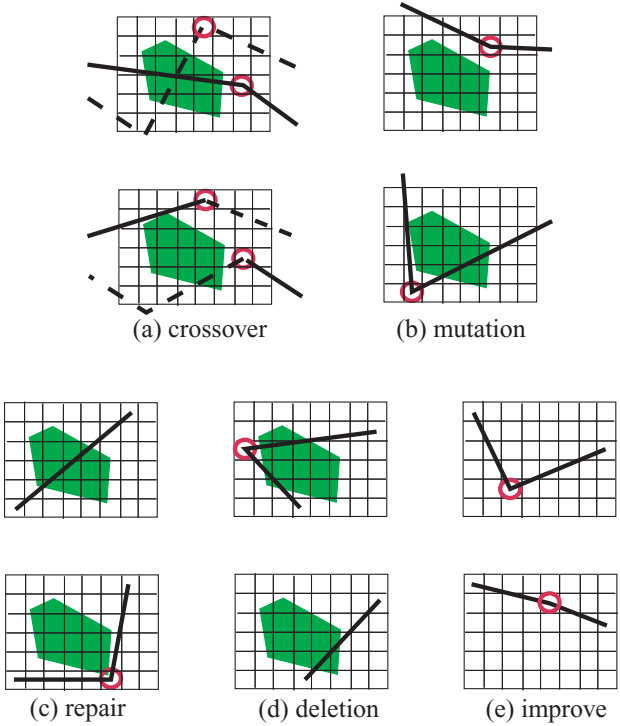}
\caption{Five specialized genetic operators that incorporates
specific knowledge of robot path planing.}
\label{f_fig4} 
\end{center}
\end{figure}

Those two commonly-used basic genetic operators, crossover and mutation, 
are not applicable for the robot path-planning problem here.  They have to be
tailored to suit for the problem and the adopted problem representation. 
In addition, to make the genetic algorithm more effective, 
three more specialized operators are designed to make use of 
available problem-specific knowledge, including knowledge of the environment 
(e.g., numbers and positions of obstacles) and the path 
(e.g., feasibility and quality of a path).  Some
of the operators combine a small-scale local search technique.  These five
operators are illustrated in Fig. \ref{f_fig4}.

\emph{Crossover} is the operator that randomly choose a node from Parent $1$
and the other node from Parent $2$.  Exchange the part after these two nodes.
Check these two offspring, and delete the part between two same nodes if it
happens. The traditional 1-point or 2-point crossover cannot be used here 
because the length of a chromosome is variable. 
The choice of different crossover sites in different parents 
can increase the variability of chromosome length,
which benefits exploration of the solution space. 

\emph{Mutation} is to randomly choose a node and replace it with a node
that is not included in the path.  Mutation is served as a key role to
diversify the solution population. Therefore, it is not necessary that a
solution is better after it is mutated.

\emph{Repair} is to repair an infeasible line segment by inserting
a suitable node between the two nodes of the segment. To locate the best
node available, a local search is applied in the neighboring girds of the
intersected obstacle. 

\emph{Deletion} is applicable for both feasible and infeasible path. Randomly
choose one node, check its two adjacent nodes and connected segments, if the
deletion of the chosen node is beneficial (turn the infeasible to the
feasible, or reduce the cost), delete the node.

\emph{Improvement} is designed for feasible solutions. Randomly chose one
node, do a local search in the neighboring grids of the node, move to 
a better location.  
This operator is used for fine tuning of a feasible solution.

These operators are necessary to evolve feasible and good quality
solutions.  The firing of these operators depends on two criteria:
probability and heuristic knowledge (e.g., if \emph{feasible} then
\emph{improvement}). In an environment with many obstacles, the portion of
feasible solutions in the initial population is small. \emph{Crossover} and
\emph{mutation} operators are far from adequate to evolve good solutions. It
is desirable to have these operators specially designed for robot path
planning, such as \emph{repair} and \emph{deletion}, 
to evolve feasible solutions from the infeasible ones. 
The important role of these problem-specific operators is 
discussed in details in Section IV.

\subsection{Outline of the Proposed Knowledge-based Genetic Algorithm}

\begin{figure}[htbp]
\begin{center}
\includegraphics[width=6cm]{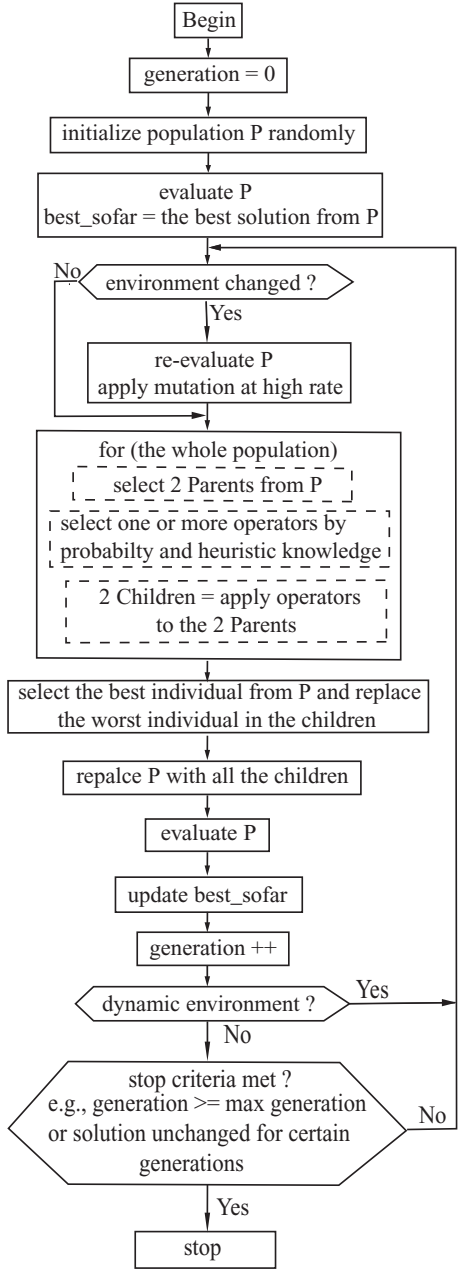}
\caption{Outline of the knowledge-based genetic algorithm 
for path planning of a mobile robot.}
\label{f_fig5} 
\end{center}
\end{figure}

An outline of the proposed knowledge-based genetic algorithm is given in Fig.
\ref{f_fig5}.  Initial solutions are generated randomly and are evaluated by
the fitness function in Eqn. (\ref{eq:fit}). The best solution from the initial population $P$ is selected and assigned to 
$best\_sofar$.
Two parents are selected
according to some selection mechanism. Here, tournament selection is used.
Then one or more genetic operators are selected and applied to the two parents 
according to some probabilities and heuristic knowledge and reproduce two children.
The selection and reproduction are applied to the whole population. The old population
consistiong of parents is called parent population $P$, and the new population consisting of
children is called child population. The whole parent population $P$ 
will be replaced by child population with elitism after application of selection and
reproduction to the whole parent population $P$. Elitism is realized by replacing the worst
individual in the child population with the best individual in the parent population $P$. 
In this way, when replacing the parent population with the child population, 
the `elite' from the parent
population $P$ won't be lost. After replacing the whole population, the child population
becomes the parent population $P$ for the next generation. 

The best solution so far ($best\_sofar$)is updated in
each generation, and it will be the final solution when a stop criterion
is satisfied. The stop criterion can either be that the preset maximum
generation is exceeded, or that the best solution remains unchanged for
certain number of generations. 

The proposed algorithm is also suitable for robot path planning in a 
dynamic environment. 
It checks the sensed environmental information either every several
generations or after a predetermined period.  If the environment is changed,
the algorithm will re-evaluate the current population according to the new
environment. At the time of change, the algorithm has gone through many
generations, the diversity of the population is relatively low, which may
make it more difficult for the GA to evolve new solutions.  To increase
diversity so as to deal with the new environment, mutation with high mutation
rate is applied to the population. From now on it starts the process to
evolve a new solution for the new environment. 
%
The proposed algorithm is computationally efficient and
would be fast enough to evolve a feasible (if not near optimal yet) 
robot path from the current robot location to the target, if the environment
is not changing too fast. In a very complicated dynamic environment, the
robot may have to wait for a feasible path evolved.

\section{Simulation Results}

To demonstrate the effectiveness of the proposed knowledge-based 
genetic algorithm, several simulations were conducted.  
In the simulations, parameters for the proposed genetic
algorithm are set as: population size is $50$, probability for mutation per
chromosome is $0.2$, and $0.9$ for all the other operators. Tournament
selection and elitism are applied. The proposed GA can deal with 
different resolutions as shown later in the discussion section. 
For simplicity, in all simulations in this section,
100 units $\times$ 100 units grids is used 
to the nodes of paths unless it is otherwise stated.
The simulation results also show that $100\times100$ resolution 
is high enough for the environments studied in this section. 
All simulations are conducted on a Pentium III PC (933 MHz) with Windows 2000
Operating System.

\subsection{Path Planning in Unstructured Environments}

\begin{figure}[htbp]
\begin{center}
\includegraphics[width=0.45\textwidth]{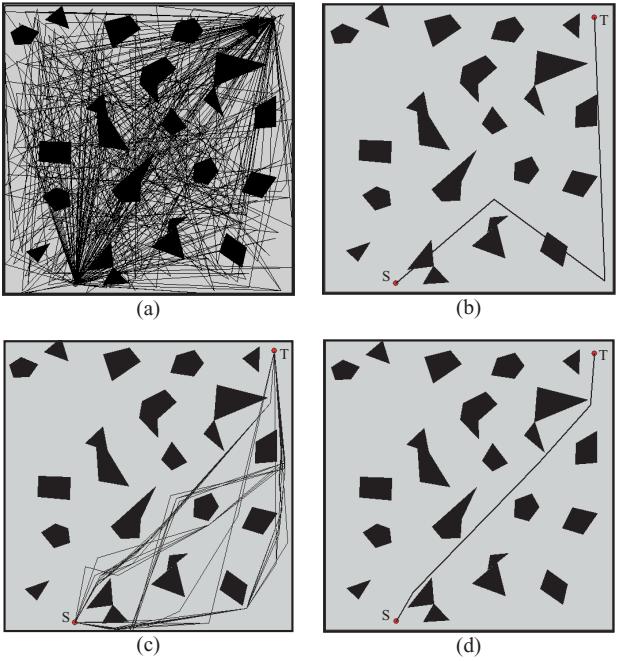}
\caption{
One typical run of robot path planning in an unstructured environment. 
(a) The randomly generated initial paths; 
(b) The best path in the initial population; 
(c) The best paths from different generations; 
(d) The final solution path.}
\label{f_fig6} 
\end{center}
\end{figure}

The proposed knowledge-based GA is first applied to an unstructured
environment with many arbitrarily shaped obstacles. 
Fig. \ref{f_fig6} shows one typical run. The algorithm
first randomly generates the initial population as shown in 
Fig. \ref{f_fig6}(a).  Fig. \ref{f_fig6}(b) displays the best solution 
in the initial population, which is even infeasible and 
has the cost of $1043.25$. Starting from this initial population,
after selection and genetic operations, generation by generation, 
the population is evolved better and better. 
Fig. \ref{f_fig6}(c) shows the best solutions obtained from several different 
generations. It shows that the best solution first becomes feasible, 
and then the genetic algorithm starts to improve the quality of solutions 
in each generation. 
The best solution in the population after certain number of generations 
is obtained as the final solution.  
Fig. \ref{f_fig6}(d) shows the finial solution obtained 
at generation $46$. 
The cost is 117.56 and the computation time is 6.09 seconds. 
To test the robustness of the proposed GA, 
it has been run for many times. For 20 runs, the average cost
is 117.91 with the standard deviation of 1.43, and 
the average computation time is 7.81 seconds with 2.58 standard deviation, 
and the average generations needed is 51 with 16 standard deviation.

\subsection{Path Planning in Complex Environments}

\begin{figure}[tb]
\begin{center}
\includegraphics[width=0.45\textwidth]{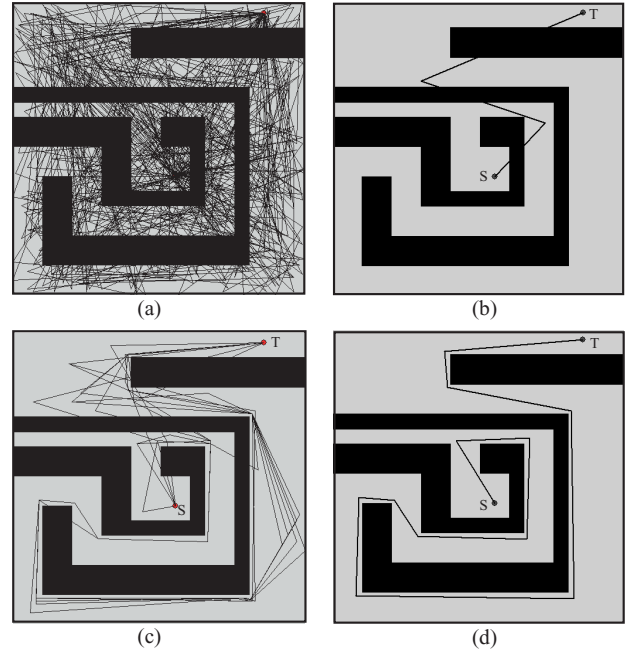}
\caption{One typical run of robot path planning 
in a maze-like environment. 
(a) The randomly generated initial paths; 
(b) The best path in the initial population; 
(c) The best paths from different generations; 
(d) The final solution.}
\label{f_fig7} 
\end{center}
\end{figure}

\begin{figure}[htbp]
\begin{center}
\includegraphics[width=0.45\textwidth]{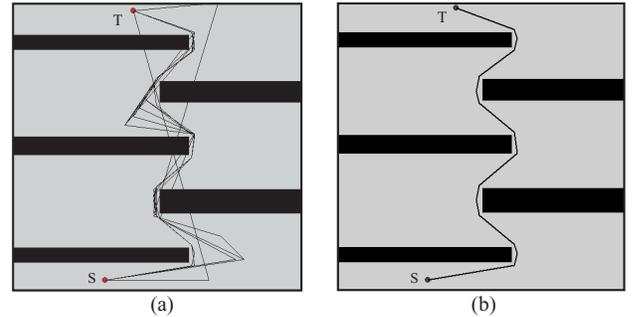}
\caption{Path planning in a zig-zag environments. 
(a) The best paths from different generations; 
(d) The final solution path.}
\label{f_fig8} 
\end{center}
\end{figure}

\begin{figure}[htbp]
\begin{center}
\includegraphics[width=0.45\textwidth]{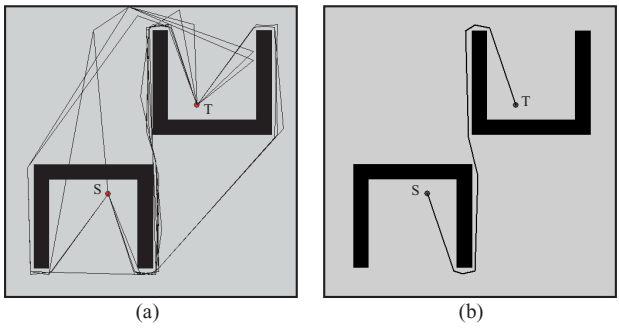}
\caption{Path planning in an environment with double U-shaped obstacles. 
(a) The best paths from different generations; 
(d) The final solution path.}
\label{f_fig9} 
\end{center}
\end{figure}

The proposed genetic algorithm can be easily applied to 
robot path planning in complicated environment, such as U-shaped or 
maze-like environment, where some potential based path
planning approaches and local path planning approaches may be trapped
\cite{Yang&Meng:2000a,Yang&Meng:tsmc2001}.  Fig. \ref{f_fig7} gives an
example to show the ability of the genetic algorithm to solve maze-like
problem.  For the environment in Fig. \ref{f_fig7}, the proposed
model is capable of obtaining the near-optimal solutions in an average of
$43$ generations for $20$ runs, where the average cost of $418.65$ (with
$4.11$ standard deviation) and computation time of $15.95$ seconds (with
$2.72$ standard deviation).  
A typical run is shown in Fig. \ref{f_fig7}, where 
Fig. \ref{f_fig7}(a) shows the initial solutions. The best solution (with a
cost of $176.96$) in the initial population is shown in Fig. \ref{f_fig7}(b), 
which is an infeasible path.  Fig. \ref{f_fig7}(c) displays the evolving best
solutions from different generations in evolution process. 
The final solution is obtained at generation $44$ with
the cost of $411.67$ and has taken $17.13$ seconds.
Figures \ref{f_fig8} and \ref{f_fig9} show two other examples of zig-zag
environment and a environment with U-shaped obstacles.  The simulation shows that the proposed GA is
able to deal with them easily.  Fig. \ref{f_fig8} shows a zig-zag
environment, where Fig. \ref{f_fig8}(a) shows the evolving solutions, and 
Fig. \ref{f_fig8}(b) shows the
final result with cost of $155.89$ obtained at generation 60 that takes
$3.43$ seconds. The average cost for 20 runs is $156.31$ with $1.99$ standard deviation,
the average generation needed is $52$with $12$ standard deviation, 
and the average time taken is $2.94$ seconds with $1.37$ standard deviation.
Fig. \ref{f_fig9} shows the simulation result in an environment with 
double U-shaped obstacles.  
Fig. \ref{f_fig9}(a) shows the results in the evolution process
while Fig. \ref{f_fig9}(b) displays the final path obtained at generation
$49$ and with cost $152.01$.  The average cost for 20 runs for this
environment is $152.44$, the average generation needed is $56$, and the
average time taken is $9.91$ seconds.

\subsection{Path Planning in a Clustered Environment}

\begin{figure}[tb]
\begin{center}
\includegraphics[width=0.45\textwidth]{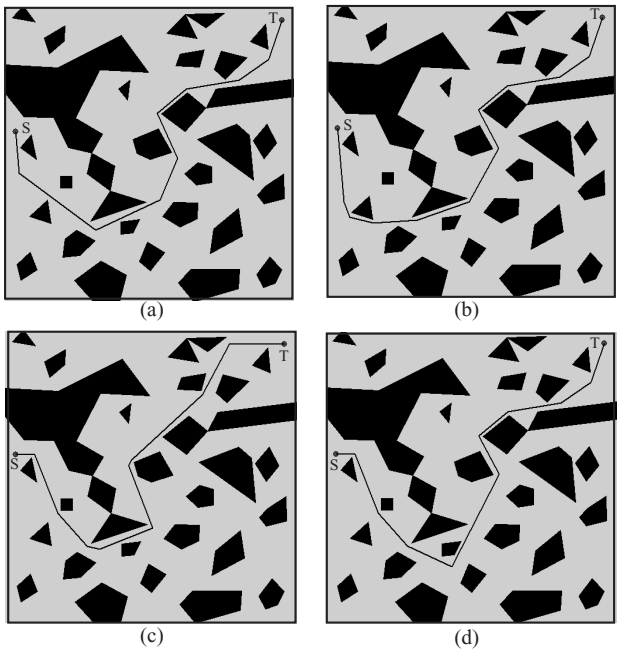}
\caption{
Path planning in a clustered environment at four different runs. 
(a) to (d): The solution paths with cost of 160.50, 160.93, 
163.95, and 170.53, respectively.}
\label{f_fig10} 
\end{center}
\end{figure}

In this simulation, the proposed genetic algorithm is applied to a 
clustered environment with many arbitrarily shaped obstacle. 
Even for obstacles with very complicated shapes, the GA can 
still accurately detect collision and evolve near-optimal solutions. 
In a very complicated environment, it is possible that the GA obtains 
different solutions in different runs 
because of the randomness involved in genetic algorithms. 
However, near-optimal solutions are guaranteed.  
Fig. \ref{f_fig10} shows four near-optimal solutions from four different runs. 
Their costs are 160.50 (a), 160.93 (b), 163.95 (c) and 170.53 (d), 
respectively, which are slightly different, 
but they are all good solution paths to this complex environment. 
The average computation time for the four runs is $20.83$ seconds.

\subsection{Path Planning in a Dynamic Environment with Moving Obstacles}

\begin{figure}[htbp]
\begin{center}
\includegraphics[width=0.45\textwidth]{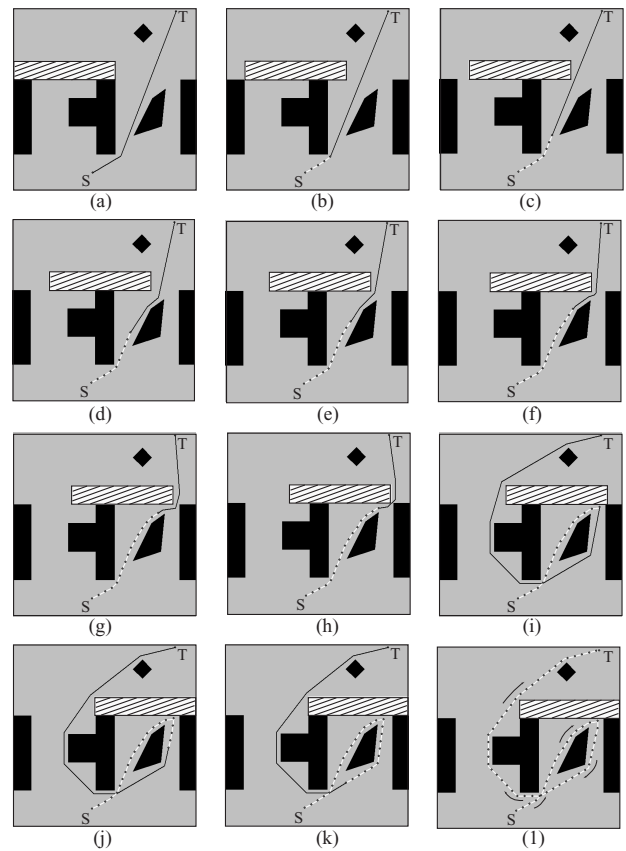} 
\caption{Path planning in a dynamic environment with a moving obstacle. 
The bar-shaped obstacle moves from the left to the right. 
(a) to (l): Snapshots at different time instances during  
the robot moves from the start location to the target location.}
\label{f_fig11}
\end{center}
\end{figure}

The proposed genetic algorithm can not only deal 
with complex static environment, 
but also be suitable for dynamic environments with 
moving obstacles while the mobile robot is moving at a certain speed. 
A simulation is presented to show the adaptability of the proposed
genetic algorithm to a changing environment with on-line planning. 
In Fig. \ref{f_fig11}, initially the slash-shaded obstacle is on the left
side, which closed the left channel for the robot. It moves from 
the left to the right like a door opening the left side 
to shut the right channel at the speed of 1 unit/second 
(with 100 unit$\times$100 unit grids applied). The robot moves at
the speed of 2 units/second. The environment information as well as the
robot position are updated every two seconds.  
Fig.s \ref{f_fig11}(a)-(g) show the snapshots of the obtained path solutions 
after updating the environment information at each time instance.  
The dots indicate the robot positions at
the time of each update. The white line shows the the trajectory of the robot
in the past, and the black line is the current path solution according to the
last updated environment information (including the 
positions of the moving obstacle and the robot).  
Fig.  \ref{f_fig11}(a) displays the obstacle original position, 
the start point of the robot, and the obtained path solution. 
In Fig. \ref{f_fig11}(b) and (c),
the moving obstacle moves to new positions, but the movement of the obstacle
has not actual impact onto the current solution. 
However, the algorithm adjusts the path slightly because of changes 
in the robot position (new start point).  
Starting from Fig. \ref{f_fig11}(d), the moving obstacle has actual
impact on the obtained best path solution.  The genetic algorithm is able to
evolve a new path to avoid collision with the moving obstacle while trying to
have the shortest path. 
As shown in Fig. \ref{f_fig11} (d)-(h), the best path
solutions remain between the gap while the door is shutting 
because they are the shortest paths with regard to robot positions 
at the moment. When the
door is shut totally, as shown in Fig. \ref{f_fig11}(i), the algorithm
obtains a complete new path immediately to escape the deadlock. When the door
is shut, the moving obstacle stops over the right side, 
but the robot position is still updated periodically, 
and the genetic algorithm obtains the best paths
according to the new start points. Obviously, only slight adjustment is
needed when necessary.

\begin{figure}[htbp]
\begin{center}
\includegraphics[width=0.45\textwidth]{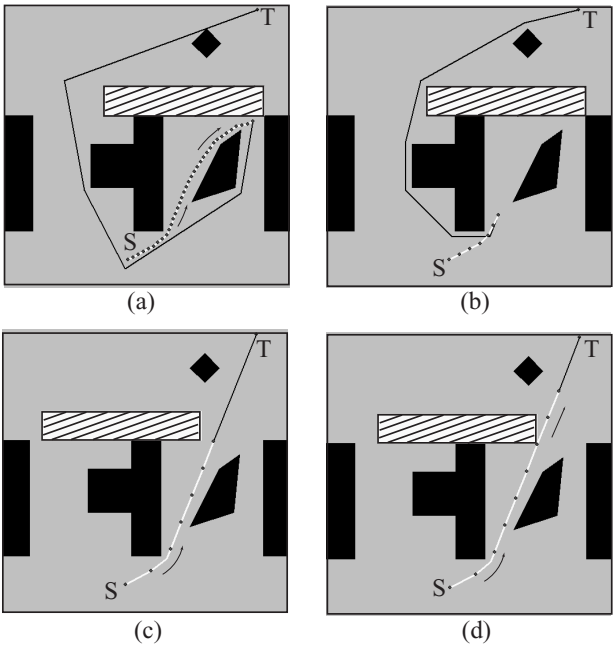} 
\caption{Path planning in the same situation as Fig. \ref{f_fig11} except
(a) with a smaller update interval at 1 second instead of 2 seconds in Fig.
\ref{f_fig11}; 
(b) with a faster obstacle moving speed at 
at 2.5 unit/second instead of 1 unit/second in Fig. \ref{f_fig11}; 
(c) and (d) with a faster robot speed at 
5 unit/second instead of 2 unit/second in Fig. \ref{f_fig11}.}
\label{f_fig12}
\end{center}
\end{figure}

The simulation shows that the genetic algorithm is able to successfully
obtain collision free and good quality solutions soon according to the
last updated environment information. Occasionally, the robot might be trapped into a certain 
location when the moving of the obstacle and moving of the robot make the robot 
switch directions along a same path. However, this should happen rarely because it needs the
right timing of moving of the obstacle and the robot. Also, randomness involved in the
genetic algorithm will reduce the likeliness of repeating the same path so as to reduce 
the chance of being trapped. 
The overall quality of solutions
in very complicated environment may possibly increase when the interval
between environmental information updates 
increases because the algorithm will have more time to refine
the solution. For example, Fig. \ref{f_fig12}(a) displays the path obtained
when the update interval is $1$ second instead of $2$ seconds in Fig.
\ref{f_fig11} at the stage of Fig. \ref{f_fig11}(i), where the quality of the
path is better. Obviously, when the moving speeds of the obstacle and the
robot are different, different path planning results will be expected. Fig.
\ref{f_fig12}(b) shows the path planning when the moving obstacle moves
faster at the speed of is 2.5 units/second while the robot moves at the same
speed as in Fig. \ref{f_fig11}. 
In Fig. \ref{f_fig12}(c) and (d), the robot moves faster (5 units/second), 
so the robot is able to pass through the right opening before it is closed by
the moving obstacle. 

\subsection{Path Planning in a Dynamic Environment with Unknown Obstacles}

\begin{figure}[htbp]
\begin{center}
\includegraphics[width=0.45\textwidth]{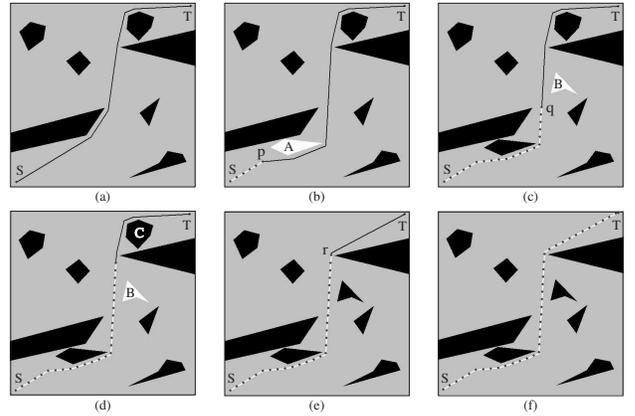}
\caption{Path Planning in a Dynamic Environment with unknown obstacles.}
\label{f_sudden} 
\end{center}
\end{figure}

In this simulation, the algorithm is applied to an environment with unknown obstacles. 
At the first, the robot environment is partially known. Unknown obstacles are sensed
when the robot is moving toward the target along the current planned path. Robot environment
with suddenly appearing obstacles is one of the dynamic environments need to be handled by 
path planning task. Fig. \ref{f_sudden} shows such a dynamic robot environment and the 
snapshots of the path planning results of one typical run. 
Fig. \ref{f_sudden}(a) shows the partially known environment at the beginning of path
planning, and the planned path with the current environment. When the robot moves to 
location $p$ along the current path, obstacle $A$ is detected and the environment information
is updated accordingly. Once the genetic algorithm senses the change, it re-evaluates the 
population, and the current best path is not feasible any more. A new path shown in
Fig. \ref{f_sudden}(b) is obtained in as short as $1.10$ seconds. The robot now
follows the new path and may adjust the path if necessary according to every new start point. 
When the robot moves to location $q$, obstacle $B$ is detected. The newly detected obstacle
has no actual impact to the best solution even though it has some impacts to some other
solutions in the population. Therefore, the robot continues moving along the unchanged path
(Fig. \ref{f_sudden}(c))
until it comes to location $r$, where obstacle $C$ is removed (Fig. \ref{f_sudden}(d) and(e)).
With the new environment
information, a new best path is evolved immediately ($0.08$ seconds). Fig. \ref{f_sudden}(f)
displays the whole path.

This simulation demonstrates that the genetic algorithm is able to react to newly 
sensed obstacles
very quickly. When the newly detected obstacle blocks the current path, it is able to 
avoid the obstacle and obtain a good quality path. When the appearance of the obstacle 
is irrelevant, the 
current path is not disturbed. The algorithm also responds to removal of an obstacle. 
The algorithm updates the start point while it moves towards the target, and the algorithm
will update the path if it finds a better path with the current start point.

\section{Discussions}

The above simulation results demonstrate that the proposed knowledge based 
GA is capable of
evolving satisfactory robot paths in complex environments. This is mainly
contributed by the specialized genetic operators that incorporate heuristic
knowledge. To show the contribution from these operators, a comparison study
between the GA with and without the specialized operators is conducted.
The comparison is conducted in the same environment as shown in 
Fig. \ref{f_fig8}. To see the performance of the GA without the developed
specialized operators, we only keep crossover and mutation operators, and
shut off all the other operators.  
As every specialized operator can be viewed
as a special mutation operator, simply shutting off those operators 
makes the two sides of the comparison have different mutation rates.  
To minimize the bad effect of this, we first set a
best mutation rate for the GA with only crossover and mutation operators.  By
running the GA for $20$ times at different mutation rates, a mutation
probability of $0.5$ is selected as the best value. Then the GAs with and
without specialized operators are run for $20$ times respectively. Statistic
analysis is shown in Table \ref{table1}.

\begin{table}[htbp]
\begin{center}
\vspace{-0.2cm}
\caption{\label{table1}
\bf Comparison of the GA with and without specialized operators 
(SD: standard deviation)}
\vspace{0.2cm}
\begin{tabular}{p{2cm}|p{2cm}||p{1.3cm}|p{1.3cm}} 
\hline
\hline
\multicolumn{2}{c||}{Specialized operators}& with & without\\
\hline
\multicolumn{2}{c||}{number of runs} & $20$ &$20$\\
\hline
\hline
Best found  & Mean & $156.31$ & $352.35$\\ \cline{2-4}
path cost  & SD & $0.47$ & $42.36$\\
\hline
Number of & Mean & $52$ & $346$\\ \cline{2-4}
generations & SD & $13$ & $84$\\
\hline
time& Mean & $2.94$ & $3.84$\\ \cline{2-4}
needed & SD & $0.77$ & $0.94$\\
\hline
\hline
\end{tabular}
\end{center}
\end{table}

\begin{figure}[htbp]
\begin{center}
\includegraphics[width=6cm]{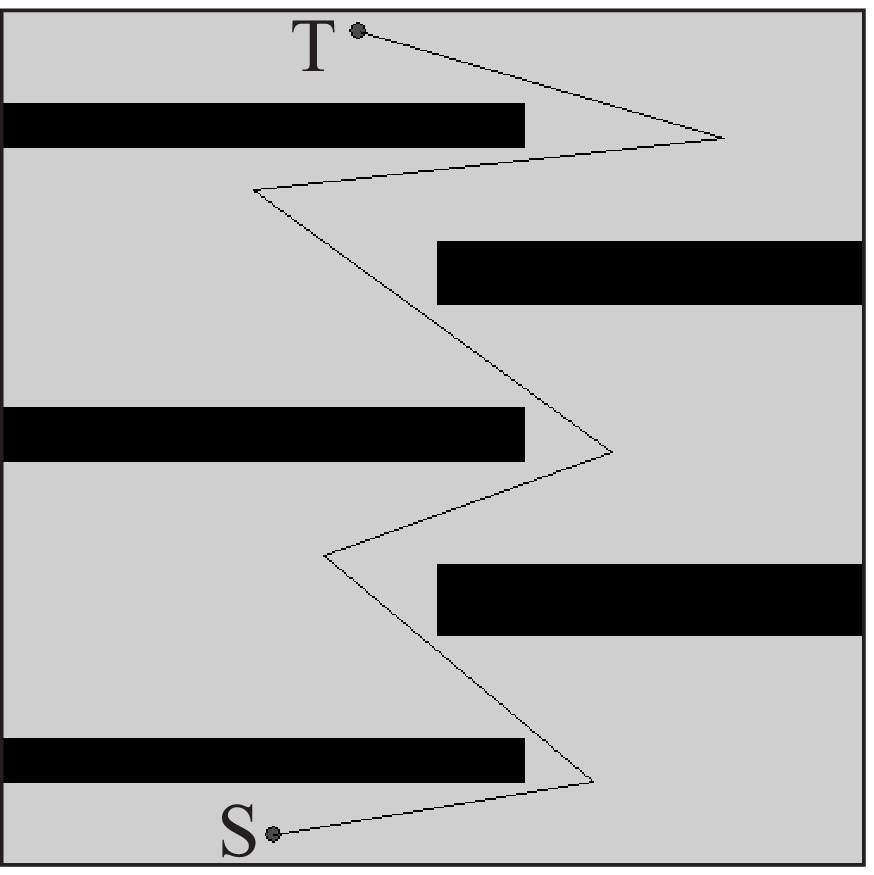} 
\caption{The path obtained by the GA with only 
the crossover and mutation operators.}
\label{f_fig13}
\end{center}
\end{figure}

Fig. \ref{f_fig13} shows the best path found in 20 runs by the GA with only
crossover and mutation operators.  The cost is $261.63$, which is found at
generation $444$ and takes $5.04$ seconds.  In comparison to Fig. \ref{f_fig8}
using the proposed approach with the specialized operators, it is clear that
without our specialized operators designed for robot path
planning, the quality of solutions deteriorates
dramatically in spite of more computation time.  
When the environment is more
complicated, the GA without the specialized operators  even cannot find a
feasible path.  For the environment in Fig. \ref{f_fig7}, a feasible path can
never be found no matter how the parameters are set. This comparison study
demonstrates that the knowledge-based genetic operators are essential for
solving the robot path planning problem.

\begin{figure}[htbp]
\begin{center}
\includegraphics[width=6cm]{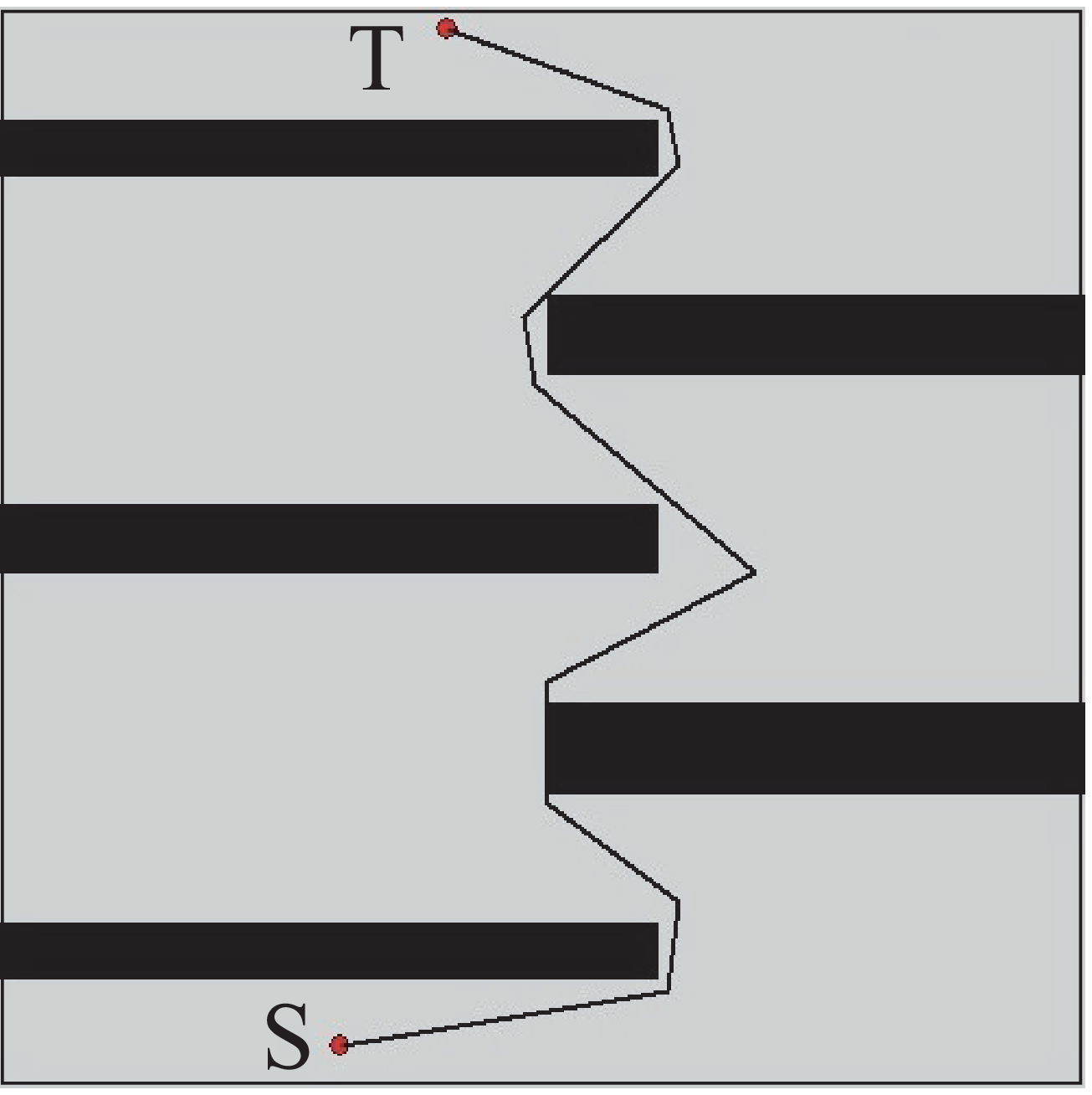} 
\caption{One typical run of the GA without crossover and mutation operators.}
\label{f_fig13+}
\end{center}
\end{figure}

The above simulation was done by shuting off repair, deletion and improve operators, 
and the simulation result proves the effectiveness of these genetic operators. 
The next question is: Are those three specialized genetic operators effective
enough to evolve good quality solutions? To answer the question, another simulation
that shuts off crossover and mutation operators is conducted. First, the simulation
is conducted with the same environment as above. Feasible solutions can be obtained,
but with much more generations and deteriorated qualities. For 20 runs, the average 
cost is $251.56$ with $24.17$ standard deviation, and $174$ generations are needed on
average. Fig. \ref{f_fig13+} shows one typical run of this simulation. The path has
the cost of $239.63$ and is obtained at generation $168$. Then the simulation
is conducted with the environment shown in Fig. \ref{f_fig7} that is much more
challenging. It turns out that the algorithm without crossover and mutation operators
fails to get a feasible path. The failure is as expected. Although the repair operator
is very powerful, yet it is not deterministic and it tries to repair one infeasible 
line segment at one time. When the path has too many infeasible line segments and the 
partially repaired paths cost more, the solution is trapped into the low cost infeasible
solutions. In this case, the only way to improve the solution is to combine other line
segments from other paths by crossover. 

The above simulations of shuting off one or more genetic operators demonstrate that
every genetic operator contributes to the success of the algorithm, and all the operators
work together to deal with different robot environments.

Simulation study also indicates that the proposed knowledge-based genetic
algorithm is practically feasible 
because the required computation time is quite reasonable. 
As introduced before, grids are used to form the intermediate
nodes of a path. Obviously, resolution is a main factor to affect the
computation time. However, simulation results show that the computation time
does not increase dramatically as the resolution increases. For example, for
the environment shown in Fig. \ref{f_fig6}, when resolution is
$100\times100$, the average computation time is $7.81$ seconds for $20$ runs.
When the resolution increases to $200\times200$, the average computation time
is $9.03$ seconds, and it takes average $12.25$ seconds to compute when the
resolution is $400\times400$.  

\begin{figure}[htbp]
\begin{center}
\includegraphics[width=0.45\textwidth]{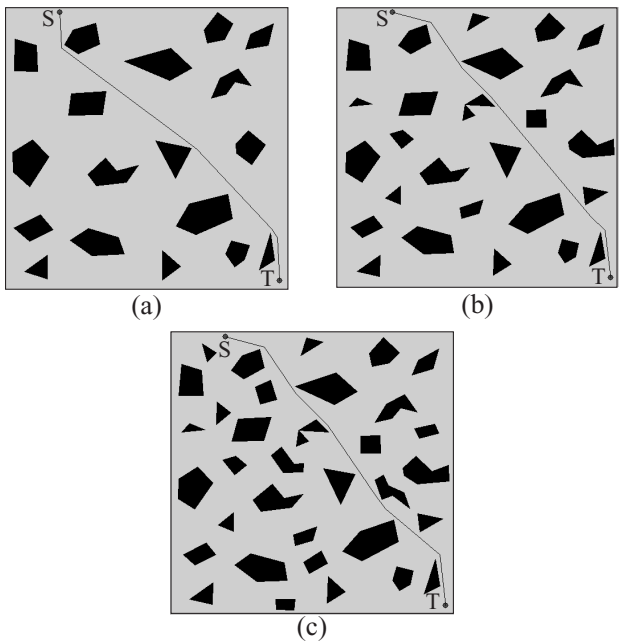}
\caption{Path planning in environments with different numbers of obstacles. 
(a) 20 obstacles, takes average 5.53 seconds;
(b) 30 obstacles, takes average 11.54 seconds;
(c) 40 obstacles, takes average 18.96 seconds.}
\label{f_fig14}
\end{center}
\end{figure}

Another factor affects the computation time is
the number of the obstacles in an environment.  Simulation study is also
conducted to show the sensitivity of the proposed GA to the number of
obstacles (a concave obstacle is considered as two or more convex obstacles).
Fig. \ref{f_fig14} shows the environment with different numbers of obstacle:
$20$, $30$ and $40$. Accordingly, the average computation time for $20$ runs
is $5.53$, $11.54$ and $18.96$ seconds, respectively. 
The simulation results shows
that the number of obstacles in the environment is the main factor affecting 
computational time of the algorithm. The reason behind this that the algorithm
needs to check collision with every obstacle. Like other genetic algorithms, 
evaluation is the bottleneck of the algorithm. 

\section{CONCLUSION}

In this paper, a knowledge-based genetic algorithm for path planning of
mobile robots is proposed.  The GA uses a simple and unique path
representation that uses natural coordinates to represent environment
and grids to form the intermediate nodes of paths. 
The classical crossover and mutation genetic operators are
tailored to the path planning problem. 
The proposed genetic algorithm also incorporates domain knowledge into 
its three problem-specific genetic operators for robot path planning. 
These operators also adopt small-scale local search based on some heuristic
knowledge. The effectiveness of these knowledge-based operators is
demonstrated by simulation studies. The simulation results also show that the
proposed genetic algorithm is applicable for both static and dynamic 
environments. The developed GA also features its one fitness function for
both feasible and infeasible solutions. 
This evaluation method accurately detects
collision between obstacles and paths, 
and effectively distinguishes qualities of both
feasible and infeasible solutions, 
which is critical for the GA to evolve better solutions. 
The capability of the propose genetic algorithm of 
dealing with moving obstacles and the efficiency 
in computation time make the proposed knowledge-based
genetic algorithm be able to be applied to real applications.

Knowledge incorporation into GAs is desirable when GAs are applied to
specific problems.  In the proposed GA, domain knowledge is
incorporated into its genetic operators. There are many 
other ways to utilize additional knowledge besides designing specialized
genetic operators. Constructing beneficial initial population is one area 
that domain knowledge could be used as guidance. In the initial population
generated randomly for robot path planning in a complex environment with many
obstacles, there are only few  or no feasible solutions. If we use domain
knowledge to generate more feasible solutions for initial populations, the GA
would work better. This would be a future work.  
Besides, the genetic algorithm includes more genetic operators, firing
of which depends on their respective probability. On-line tuning of these
probabilities would be much desirable, which would another important future work.


\bibliographystyle{ieeetr}
\bibliography{planning}

\begin{thebibliography}{10}

\bibitem{Theocharous&Mahadevan:2002}
G.~Theocharous and S.~Mahadevan, ``Approximate planning with hierarchical
  partially observable markov decision process models for robot navigation,''
  in {\em Proc. of IEEE Intl. Conf. on Robotics and Automation},
  pp.~1347--1352, May 2002.

\bibitem{Laroche&etal:1999}
P.~Laroche, F.~Charpillet, and R.~Schott, ``Mobile robotics planning using
  abstract markov decision process,'' in {\em IEEE International Conference on
  Tools with Artificial Intelligence}, pp.~299--306, 1999.

\bibitem{Laroche:2000}
P.~Laroche, ``Building efficient partial plans using markov decision process,''
  in {\em 12th IEEE International Conference on Tools with Artificial
  Intelligence}, pp.~156--163, Nov 2000.

\bibitem{Cassandra&etal:1996}
A.~R. Cassandra, L.~P. Kaelbling, and J.~A. Kurien, ``Acting under uncertainty:
  discrete bayesian models for mobile-robot navigation,'' in {\em Proceedings
  of IROS}, pp.~963--972, 1996.

\bibitem{Latombe:1991}
J.~C. Latombe, {\em Robot motion planning}.
\newblock Boston: Kluwer Academic Publisher, 1991.

\bibitem{Lozano-Perez:1983}
T.~Lozano-Perez, ``Spatial planning: A configuration approach,'' {\em IEEE
  Trans. on Computers}, vol.~C-32, pp.~108--120, Feb 1983.

\bibitem{Sharir:1989}
M.~Sharir, ``Algorithmic motion planning in robotics,'' {\em Computer},
  vol.~22, pp.~9--19, March 1989.

\bibitem{Khosla&Volpe:1988}
P.~Khosla and R.~Volpe, ``Superquadric artificial potentials for obstacle
  avoidance and approach,'' in {\em Proc. of IEEE Intl. Conf. on Robotics and
  Automation}, vol.~3, (Philadelphia, PA, USA), pp.~1778--1784, April 1988.

\bibitem{Rimon&Doditschek:1992}
E.~Rimon and D.~E. Doditschek, ``Exact robot navigation using artificial
  potential fields,'' {\em IEEE Trans. on Robotics and Automation}, vol.~8,
  pp.~501--518, 1992.

\bibitem{Yang&Meng:2000a}
S.~X. Yang and M.~Meng, ``An efficient neural network approach to dynamic robot
  motion planning,'' {\em Nerual Networks}, vol.~13, pp.~143--148, 2000.

\bibitem{Yang&Meng:2000b}
S.~X. Yang and M.~Meng, ``An efficient neural network method for real-time
  motion planning with safety consideration,'' {\em Robotics and Autonomous
  Systems}, vol.~32, pp.~115--128, 2000.

\bibitem{Yang&Meng:tsmc2001}
S.~X. Yang and M.~Meng, ``Neural network approaches to dynamic collision-free
  robot trajectory generation,'' {\em IEEE Trans. on Systems, Man, and
  Cybernetics, Part B}, vol.~31, pp.~302--318, June 2001.

\bibitem{Yang&Meng:tnninpress}
S.~X. Yang and M.~Q.-H. Meng, ``Real-time collision-free motion planning of
  mobile robots using neural dynamics based approaches,'' {\em IEEE Trans. on
  Neural Networks}.
\newblock To apear in Nov. 2003.

\bibitem{Holland:1975}
J.~Holland, {\em Adaptation in natural and artificial systems}.
\newblock Ann Arbor: University of Michigan Press, 1975.

\bibitem{Davis:1987}
L.~D. (Editor), {\em Genetic algorithms and simulated annealing}.
\newblock Los Altos, California: Morgan Kaufman Publishers, 1987.

\bibitem{Michalewicz:1994}
Z.~Michalewicz, {\em Genetic algorithms + Data structures = Evolution programs,
  2nd extended edition}.
\newblock Springer-Verlag, 1994.

\bibitem{Sugihara&Smith:1997a}
K.~Sugihara and J.~Smith, ``Genetic algorithms for adaptive motion planning of
  an autonomous mobile robot,'' in {\em Proc. of IEEE Intl. Symposium on
  Computational Intelligence in Robotics and Automation}, pp.~138--143, 7 1997.

\bibitem{Sugihara&Smith:1997b}
K.~Sugihara and J.~Smith, ``Genetic algorithms for adaptive planning of path
  and trajectory of a mobile robot in 2d terrains,'' Tech. Rep. ICS-TR-97-04,
  Department of Information and Computer Sciences, Univesity of Hawaii, Hawii,
  USA, May 1997.

\bibitem{Sugihara&Smith:1997c}
K.~Sugihara and J.~Smith, ``Ga-based motion planning for underwater robot
  vehicles,'' in {\em Proc. of 10th Intl. Symposium on Unmanned Untethered
  Submersible Technology}, pp.~406--416, 9 1997.

\bibitem{Ashiru&etal:1996}
I.~Ashiru, C.~Czarnecki, and T.~Routen, ``Characteristics of a genetic based
  approach to path planning for mobile robots,'' {\em J. Network and Computer
  Applications}, vol.~19, pp.~149--169, 1996.

\bibitem{Gallardo&etal:1998}
D.~Gallardo, O.~Colomina, F.~Florez, and R.~Rize, ``A genetic algorithm for
  robust motion planning,'' in {\em 11th Intl. Conf. on Industrial and
  Engineering Applications of Artificial Intelligence and Expert Systems},
  vol.~2, (Benicassim, Spain), pp.~115--121, June 1998.

\bibitem{Tu&Yang:2003}
J.~Tu and S.~X. Yang, ``Genetic algorithm based path planning for a mobile
  robot,'' in {\em Proc. of IEEE Intl. Conf. on Robotics and Automation},
  (Taiwan), September 2003.
\newblock accepted.

\bibitem{Grefenstette:1987}
J.~J. Grefenstette, ``Incorporating problem specific knowledge into genetic
  algorithms,'' in {\em Genetic Algorithm and Simulated Annealing} (L.~Davis,
  ed.), pp.~42--60, Los Altos, California: Morgan Kaufman Publishers, 1987.

\bibitem{Blum&Eskandarian:2002}
J.~Blum and A.~Eskandarian, ``Domain-specific genetic agents for flow
  optimization of freight railroad traffic,'' in {\em Computers in Railways:
  Proc. of 8th Intl. Conf. on Computer Aided Design, Manufacturing and
  Operation in the Railway and Other Advanced Mass Transit Systems}, (Lemnos,
  Greece), pp.~787--796, June 2002.

\bibitem{Janikow&Michalewicz:1990}
C.~Janikow and Z.~Michalewicz, ``A specialized genetic algorithm for numerical
  optimization problems,'' in {\em Proc. of the 2nd Intl. Conf. on Tools for
  Artificial Intelligence}, (Herndon, VA, USA), pp.~798--804, Nov. 1990.

\bibitem{Suh&Gucht:1987}
J.~Y. Suh and D.~V. Gucht, ``Incorporating heuristic information into genetic
  search,'' in {\em Genetic Algorithms and Their Applications: Proc. of the 2nd
  Intl. Conf. on Genetic Algorithms}, (Cambridge, MA, USA), pp.~100--107, July
  1987.

\bibitem{Akbarzadeh-T&Jamshidi:1997}
M.~Akbarzadeh-T and M.~Jamshidi, ``Incorporating a-priori expert knowledge in
  genetic algorithms,'' in {\em Proc. of IEEE Conf. on Computational
  Intelligence in Robotics and Automation}, (Monterey, California, USA),
  pp.~300--305, 7 1997.

\bibitem{Janikow:1993}
C.~Janikow, ``A knowledge-intensive genetic algorithm for supervised
  learning,'' {\em Machine Learning}, vol.~13, pp.~189--228, 1993.

\bibitem{Tripathi&etal:1996}
A.~Tripathi, D.~Vidyarthi, and A.~Mantri, ``A genetic task algorithm for
  distribution computing systems incorporating problem specific knowledge,''
  {\em Intl. J. of High Speed Computing}, vol.~8, no.~4, pp.~363--370, 1996.

\bibitem{Ramsey&Grefenstette:1993}
C.~Ramsey and J.~Grefenstette, ``Case-based initialization of genetic
  algorithms,'' in {\em Proc. of 5th Intl. Conf. on Genetic Algorithms},
  pp.~84--91, 1993.

\bibitem{Andres&etal:2000}
B.~D.~A. Toro, J.~Giron-Sierra, P.~Fernandez-Blanco, J.~D.~L. Cruz, and
  J.~Lopez-Orozco, ``Parallel genetic algorithms with a continuity operator
  that allows for knowledge inclusion,'' in {\em Proc. of the 2000 Congress on
  Evolutionary Computation}, vol.~2, (: La Jolla, CA, USA), pp.~1113--1137,
  July 2000.

\bibitem{Louis&Zhao:1995}
S.~Louis and F.~Zhao, ``Domain knowledge for genetic algorithms,'' {\em Intl.
  J. of Expert Systems}, vol.~8, no.~3, pp.~195--211, 1995.

\bibitem{Antonisse&Keller:1987}
H.~Antonisse and K.~Keller, ``Genetic operators for high-level knowledge
  representation,'' in {\em Genetic Algorithms and Their Applications: Proc. of
  the 2nd Intl. Conf. on Genetic Algorithms}, (Cambridge, MA, USA), pp.~69--76,
  July 1987.

\bibitem{Davis&Coombs:1987}
L.~Davis and S.~Coombs, ``Genetic algorithms and communication link speed
  design: constrains and operators,'' in {\em Genetic Algorithms and Their
  Applications: Proc. of the 2nd Intl. Conf. on Genetic Algorithms},
  (Cambridge, MA, USA), pp.~257--260, July 1987.

\bibitem{Coombs&Davis:1987}
S.~Coombs and L.~Davis, ``Genetic algorithms and communication link speed
  design: theoretical considerations,'' in {\em Genetic Algorithms and Their
  Applications: Proc. of the 2nd Intl. Conf. on Genetic Algorithms},
  (Cambridge, MA, USA), pp.~252--256, July 1987.

\bibitem{Shibata&Fukuda:1993}
T.~Shibata and T.~Fukuda, ``Intelligent motion planning by genetic algorithm
  with fuzzy critic,'' in {\em Proc. of Intl. Symposium on Intelligent
  Control}, pp.~565--570, Aug. 1993.

\bibitem{Shibata&etal:1992}
T.~Shibata, T.~Fukuda, K.~Kosuge, and F.~Arai, ``Selfish and coordinative
  planning for multiple mobile robots by genetic algorithm,'' in {\em Proc. of
  31th Conf. on Decision and Control}, pp.~2686--2691, 12 1992.

\bibitem{Habib&Asama:1991}
M.~K. Habib and H.~Asama, ``Effecient method to generate collision free paths
  for autonomous mobile robot based on new free space structuring approach,''
  in {\em Proc. of IEEE/RSJ Intl. Workshop on Intelligent Robotics and Systems
  IROS'91}, vol.~2, pp.~563--567, 1991.

\bibitem{Hocaoglu&Sanderson:2001}
C.~Hocaoglu and C.~Sanderson, ``Planning multiple paths with evolutionary
  speciation,'' {\em IEEE Trans. on Evolutionary Computation}, vol.~5,
  pp.~169--191, 6 2001.

\bibitem{Xiao&etal:1997}
J.~Xiao, Z.~Michalewicz, L.~Zhang, and K.~Trojanowski, ``Adaptive evolutionary
  planner/navigator for mobile robots,'' {\em IEEE Trans. on Evolutionary
  Computation}, vol.~1, pp.~18--28, 4 1997.

\bibitem{Trojanowski&etal:1997}
K.~Trojanowski, Z.~Michalewicz, and J.~Xiao, ``Adding memory to the
  evolutionary planner/navigator for mobile robots,'' in {\em IEEE Intl. Conf.
  on Evolutionary Computation}, (Indianapolis), pp.~483--487, 4 1997.

\bibitem{Xiao&etal:1996}
J.~Xiao, Z.~Michalewicz, and L.~Zhang, ``Evolutionary planner/navigator:
  operator performance and self-tuning,'' in {\em IEEE Intl. Conf. on
  Evolutionary Computation}, (Nagoya, Japan), pp.~366--371, 5 1996.

\bibitem{Xiao:1997}
J.~Xiao, ``The evolutionary planner/navigator in a mobile robot environment,''
  in {\em Handbok of Evolutionary Computation, release 1997} (T.~Back,
  D.~Fogel, and Z.~Michalewicz, eds.), p.~G3.6, IOP Publishing Ltd., 1997.

\bibitem{Hocaoglu&Sanderson:1995}
C.~Hocaoglu and A.~C. Sanderson, ``Evolutionary supposition using minimal
  representation size clustering,'' in {\em Evolutionary Programming IV. Proc.
  of the 4th Annual Conf. on Evolutionary Programming}, (San Diego, CA, USA),
  pp.~187--203, March 1995.

\bibitem{Hocaoglu&Sanderson:1997}
C.~Hocaoglu and A.~C. Sanderson, ``Multimodal function optimization using
  minimal representation size clustering and its application to planning
  multipaths,'' {\em J. of Evolutionary Computation}, vol.~5, no.~1,
  pp.~81--104, 1997.

\bibitem{Pavlidis:1982}
T.~Pavlidis, {\em Graphic and immage processing}.
\newblock Computer Science Press, 1982.

\bibitem{Cohen&etal:1995}
J.~D. Cohen, M.~C. Lin, D.~Manocha, and M.~K. Ponamgi, ``I-collide: An
  interactive and exact collision detection system for large-scale
  environments,'' in {\em Proceedings of ACM, International 3D Graphics
  Conference}, pp.~189--196, October 1995.

\end{thebibliography}

\end{document}